\documentclass[11pt]{article}

\usepackage[preprint]{acl}
\usepackage{times}
\usepackage{latexsym}
\usepackage[T1]{fontenc}
\usepackage[utf8]{inputenc}
\usepackage{microtype}
\usepackage{inconsolata}
\usepackage{graphicx}
\usepackage{booktabs}
\usepackage{amsmath}
\usepackage{algorithm}
\usepackage{algpseudocode}
\usepackage{makecell}

\newcommand{\methodname}{Chunk-Level Guided Generation}

\newcommand{\lgsfull}{Likelihood-Guided Selection (LGS)}
\newcommand{\cgsfull}{Contrastive-Guided Selection (CGS)}

\newcommand{\lgs}{Likelihood-Guided Selection}
\newcommand{\cgs}{Contrastive-Guided Selection}

\newcommand{\lgsshort}{LGS}
\newcommand{\cgsshort}{CGS}

\title{Off-the-Shelf LLMs as Process Scorers: Training-Free Alternative to PRMs for Mathematical Reasoning}

\author{
  Atoosa Chegini\textsuperscript{1}, Soheil Feizi\textsuperscript{1} \\
  \textsuperscript{1}Department of Computer Science, University of Maryland \\
  {\small \textbf{Correspondence:} \href{mailto:atoocheg@umd.edu}{atoocheg@umd.edu}} \\
}

\begin{document}
\maketitle

\begin{abstract}
Selecting the best response from multiple small-model samples using a stronger 
scorer is a simple inference-time strategy, but fails in practice because the 
small model has already committed to diverging reasoning paths that no scorer 
can correct; PRM guided search avoids this by scoring candidate continuations 
during generation, but requires a reward model trained with step-level labels.

We propose \emph{\methodname}, a training-free alternative that uses an off-the-shelf large language model as a process scorer: a small model samples $k$ fixed-length candidate chunks at each step, while the larger language model scores the candidates using likelihoods, without generating any text.
The selected chunk is committed before the next step, steering generation before errors can propagate.

We instantiate this framework with two selection rules:
\emph{\lgsfull}, which selects the chunk with the highest length-normalized
large-model log-probability, and \emph{\cgsfull}, which subtracts the small
model's log-probability to favor chunks where the large model's preference
diverges from the small model's.
We show that scoring variable-length reasoning steps with large-model likelihoods is unreliable due to a systematic length bias that persists even after length normalization, and that fixed-length chunks avoid this confound. 

On GSM8K, MATH, Minerva Math, AMC23, and AIME24 with Qwen2.5-1.5B guided by 
Qwen2.5-32B and Llama-3.2-1B guided by Llama-3.1-70B, \cgsshort\ outperforms 
majority voting by up to 28~pp (e.g., 83.9\% vs.\ 56.0\% on GSM8K with the 
Llama pair at $k{=}32$) and, under matched guidance budgets, matches or 
outperforms Qwen2.5-Math-PRM-72B guided search on most benchmarks without any 
reward-model training (e.g., 39.0\% vs.\ 32.4\% on Minerva Math with the Qwen 
pair at $k{=}32$).
With Qwen2.5-7B guided by Qwen2.5-72B, \cgsshort\ reaches 81.8\% on MATH and 
63.6\% on Minerva Math at $k{=}16$, surpassing majority voting by 4--6~pp.

Finally, we show that \methodname\ produces substantially shorter reasoning traces than PRM guided search, with PRM responses up to 80\% longer on GSM8K.

\end{abstract}

\section{Introduction}

Large language models have made substantial progress on mathematical reasoning, but using a large model for every inference query remains expensive. A practical alternative is to sample multiple solutions from a smaller model and then select the best one. Majority voting \citep{wang2023selfconsistency}, verifier-based Best-of-$N$ \citep{cobbe2021gsm8k}, and self-certainty-based selection \citep{kang2025selfcertainty} follow this post-hoc paradigm: the small model first generates complete candidate solutions, and selection happens only after generation has finished.

These post-hoc methods are simple, but they have an important limitation.
Small-model samples can diverge early into different incorrect reasoning paths. Once every candidate has already committed to flawed intermediate steps, even a strong scorer can only choose among completed trajectories; it cannot intervene before the error propagates.

Process reward model (PRM) guided search \citep{she2025rprm,snell2024scalingllmtesttime} addresses this timing problem by moving selection inside generation. Instead of scoring only completed responses, it repeatedly samples candidate reasoning steps, scores them with a trained PRM, and appends the highest-scoring step before continuing generation. This generate-score-select loop can guide intermediate reasoning decisions, but it requires a reward model trained for step-level evaluation \citep{lightman2023letsverifystepbystep}, such as Qwen2.5-Math-PRM-72B \citep{zhang2025lessons}.

We propose \emph{\methodname}, a training-free alternative that keeps the
generation-time selection structure of PRM guided search but removes the need
for a trained reward model. At each step, a small model samples $k$
fixed-length candidate chunks of length $L$. A larger language model scores candidate chunks using likelihoods, and the highest-scoring chunk is appended to the context before generation continues. The large model never generates text for the final answer; it only scores candidate continuations produced by the small model, which can be evaluated in parallel.

We study two scoring rules. \emph{\lgsfull} selects the chunk with the
highest length-normalized large-model log-probability. \emph{\cgsfull}
subtracts the small model's likelihood from the large model's likelihood,
favoring chunks that the larger model finds more plausible than the small
model. This emphasizes continuations where the larger model contributes
information beyond the small model's own preferences.

A natural question is whether large-model likelihoods are already sufficient for post-hoc selection. We show they are not: Best-of-$N$ with large-model likelihood scoring often underperforms majority voting on MATH and AMC23, and can even degrade as $k$ grows (55.8\% → 52.6\% on MATH with Qwen2.5-1.5B guided by Qwen2.5-32B). The failure is not the scorer — the same likelihood signal drives substantial gains at chunk level. The bottleneck is timing.

A key design choice is to guide generation using fixed-length chunks rather than naturally delimited reasoning steps. We show that large-model likelihoods are biased toward longer reasoning steps even after length normalization. Fixed-length chunks avoid this confound because all candidates at a given decision point have the same length, making their scores directly comparable.

We evaluate \emph{\methodname} on five mathematical reasoning benchmarks: GSM8K \citep{cobbe2021gsm8k}, MATH \citep{hendrycks2021math}, Minerva Math \citep{lewkowycz2022minerva}, AMC23\footnote{\url{https://huggingface.co/datasets/math-ai/amc23}}, and AIME24\footnote{\url{https://huggingface.co/datasets/HuggingFaceH4/aime_2024}}. Across Qwen2.5-1.5B guided by Qwen2.5-32B \citep{qwen25} and Llama-3.2-1B guided by Llama-3.1-70B \citep{llama3}, chunk-level guidance substantially outperforms post-hoc selection methods, including Majority@$k$ \citep{wang2023selfconsistency}, Self-Certainty and Borda count \citep{kang2025selfcertainty}, and Best-of-$N$ \citep{cobbe2021gsm8k}. Under matched guidance budgets, our \emph{\cgs} matches or outperforms PRM guided search using Qwen2.5-Math-PRM-72B \citep{qwen25math,zhang2025lessons} on most benchmarks, without using any reward model.

The gains also persist when the generator is scaled to Qwen2.5-7B \citep{qwen25} guided by Qwen2.5-72B. \cgsshort\ achieves the best average accuracy among methods that do not use a trained reward model, outperforming the strongest post-hoc baseline on average across five benchmarks. It comes within 0.4~pp of the standalone 72B greedy baseline and within 1.5~pp of PRM-72B guided search on average, while requiring no reward-model training.

Finally, we find that these gains do not come from generating longer solutions:
\cgsshort\ produces substantially shorter reasoning traces than PRM guided search,
with PRM responses containing up to 80\% more tokens on GSM8K.

Our contributions are:
\begin{itemize}

\item We show that an off-the-shelf large language model, used only as a likelihood scorer and without any reward-model training, is an effective scorer for generation-time guidance. We instantiate this idea with two selection rules:
\lgs{} ($\ell_{\pi_l}$) and \cgs{} ($\ell_{\pi_l} - \ell_{\pi_s}$).

\item We show that large-model log-probabilities are unreliable for scoring variable-length reasoning steps due to a systematic length bias that persists even after length normalization, and that fixed-length chunks resolve this.

\item Our results show consistent gains over post-hoc selection methods across three model pairs: Qwen2.5-1.5B guided by 32B, Llama-3.2-1B guided by 70B, and Qwen2.5-7B guided by 72B. \cgsshort\ matches or surpasses PRM-72B guided search in the two smaller-generator settings, and comes within 1.5~pp of PRM-72B on average in the 7B setting, while requiring no reward-model training.

\item \methodname\ produces substantially shorter reasoning traces than PRM guided search, with PRM responses averaging 80\% more tokens on GSM8K.

\end{itemize}

\newpage
\section{Related Work}
\label{sec:related}

\paragraph{Sampling and post-hoc selection.}
Self-consistency \citep{wang2023selfconsistency} samples multiple chain-of-thought solutions and takes a majority vote over their final answers.
Verifier-based Best-of-$N$ methods \citep{cobbe2021gsm8k} generate complete responses, score them with a verifier, and select the highest-ranked response.
\citet{snell2024scalingllmtesttime} show that Best-of-$N$ becomes increasingly suboptimal on harder problems as test-time compute grows. \citet{kang2025selfcertainty} propose self-certainty as a training-free confidence signal for Best-of-$N$ selection.
These methods operate \emph{post-hoc}: generation is complete before selection, so early reasoning errors can only be avoided if another sampled trajectory corrects them.

\paragraph{Trained scorers and step-level guidance.}
Process reward models (PRMs) are trained to assign scores to individual steps in a reasoning trajectory, using either human annotations \citep{lightman2023letsverifystepbystep,uesato2022solving} or automated rollout- or search-based labels \citep{wang2024mathshepherd,luo2024omegaprm}. \citet{yuan2024free} reduces this supervision cost by deriving PRM-like signals from outcome-supervised reward models, avoiding explicit step-level labels while still requiring reward-model training. 
At inference time, \citet{she2025rprm} propose a guided search procedure that generates $N$ candidate continuations for the next reasoning step, scores each with a trained PRM, and appends the best; \citet{khalifa2023grace} follow a 
similar generate-score-select structure but use a discriminator trained to 
distinguish correct from incorrect reasoning steps.
Unlike these approaches, our method requires no reward-model training and uses an off-the-shelf large language model to score fixed-length chunks of $L$ tokens, rather than variable-length reasoning steps; we compare directly against Qwen2.5-Math-PRM-72B \citep{zhang2025lessons}.

\paragraph{Large-to-small collaborative generation.}
Recent training-free methods improve small-model reasoning by having a larger model contribute text at inference time. Speculative Thinking \citep{yang2025speculativethinking} delegates selected reflective reasoning steps to a larger model; MentorCollab \citep{liu2026mentorcollab} injects 
large-model lookahead segments when a verifier predicts guidance is useful; and Tandem \citep{tandem2026} primes the small model with compact reasoning insights generated by the large model upfront. In contrast, our large model contributes no generated text: the small model is the sole generator, and the large model only scores fixed-length chunks already sampled from it.

\paragraph{Contrastive and speculative decoding.}
Contrastive decoding \citep{li2023contrastivedecoding,obrien2023contrastivedecoding} 
improves text quality and reasoning by modifying the decoding distribution at each token step of a single generation, upweighting tokens the large model favors over the small model using $\log \pi_\text{large}(x) - \log \pi_\text{small}(x)$ and suppressing tokens both models agree on; our \cgsshort\ adapts this signal to chunk-level selection among $k$ small-model candidates rather than token-level steering.
Classical speculative decoding \citep{leviathan2023fast,chen2023accelerating} uses a small draft model to propose tokens that a larger target model verifies in parallel, reducing latency while preserving the target model's output distribution.
RSD \citep{liao2025rsd} incorporates a process reward model to evaluate draft tokens and dynamically decide whether to invoke the target model, biasing selection toward higher-reward outputs; SPECS \citep{cemri2025specs} generates candidate reasoning traces with a draft model and scores them using both the target model and a reward model to improve accuracy at test time.

\section{Method}
\label{sec:method}

Let $\pi_s$ and $\pi_l$ denote the small and large language models.
Given a question $q$, let $h$ denote the current generation context, initialized as $h=q$.
At each iteration, the small model samples $k$ candidate continuations of length $L$
tokens, $c_1,\ldots,c_k$, autoregressively from $\pi_s(\cdot \mid h)$, where $L$ is a
hyperparameter controlling chunk length.
For a candidate chunk $c=(c^1,\ldots,c^{|c|})$ where $|c|=L$ unless an end-of-sequence token is produced, in which case $|c| \leq L$, we define its length-normalized log-probability under model $\pi$ as
\begin{equation}
  \ell_\pi(c \mid h) =
  \frac{1}{|c|} \sum_{j=1}^{|c|} \log \pi(c^j \mid h, c^{<j}).
  \label{eq:logp}
\end{equation}

    We consider two scoring functions. \lgs\ scores using only the large model's likelihood:
\begin{equation}
  S_{\mathrm{LGS}}(c_i) = \ell_{\pi_l}(c_i \mid h).
  \label{eq:guided}
\end{equation}

\cgs\ scores by subtracting the small model's likelihood:
\begin{equation}
  S_{\mathrm{CGS}}(c_i) = \ell_{\pi_l}(c_i \mid h) - \ell_{\pi_s}(c_i \mid h).
  \label{eq:contrastive}
\end{equation}
At each iteration, we append the highest-scoring chunk to the context,
\begin{equation}
  h \leftarrow (h, c_{i^\star}),
  \qquad
  i^\star = \arg\max_{i \in \{1,\ldots,k\}} S(c_i),
\end{equation}
where $S$ is either $S_{\mathrm{LGS}}$ or $S_{\mathrm{CGS}}$. We repeat until the selected chunk contains an end-of-sequence token.

\begin{table*}[!t]
  \centering
  \setlength{\tabcolsep}{4pt}
  \resizebox{\textwidth}{!}{%
  \begin{tabular}{lrrrrrrrrrrrrrrr}
  \toprule
  & \multicolumn{3}{c}{\textbf{GSM8K}} & \multicolumn{3}{c}{\textbf{MATH}} & \multicolumn{3}{c}{\textbf{Minerva}} & \multicolumn{3}{c}{\textbf{AMC23}} & \multicolumn{3}{c}{\textbf{AIME24}} \\
  \cmidrule(lr){2-4}\cmidrule(lr){5-7}\cmidrule(lr){8-10}\cmidrule(lr){11-13}\cmidrule(lr){14-16}
  \textbf{Method} & $k{=}8$ & $k{=}16$ & $k{=}32$ & $k{=}8$ & $k{=}16$ & $k{=}32$ & $k{=}8$ & $k{=}16$ & $k{=}32$ & $k{=}8$ & $k{=}16$ & $k{=}32$ & $k{=}8$ & $k{=}16$ & $k{=}32$ \\
  \midrule
  \multicolumn{16}{l}{\textit{Qwen2.5-1.5B $\to$ Qwen2.5-32B}} \\
  \midrule
    PRM guided search  & \textbf{86.0} & 89.2 & 92.3 & \textbf{64.2} & \textbf{66.4} & \textbf{69.8} & 25.0 & 28.3 & 32.4 & 36.7 & 39.2 & 43.3 & 5.6 & 6.7 & \textbf{11.1} \\
  \lgsshort\ (ours)             & 85.4 & 87.6 & 90.4 & 60.4 & 62.2 & 68.2 & 31.6 & 33.8 & 36.0 & 35.8 & 39.2 & 43.3 & \textbf{6.7} & 7.8 & \textbf{11.1}\\
  \cgsshort\ (ours)        & 85.7 & \textbf{89.3} & \textbf{92.5} & \textbf{64.2} & 66.2 & 69.6 & \textbf{32.7} & \textbf{36.8} & \textbf{39.0} & \textbf{38.3} & \textbf{40.8} & \textbf{46.7} & \textbf{6.7} & \textbf{8.9} & \textbf{11.1} \\
  \midrule
  \multicolumn{16}{l}{\textit{Llama-3.2-1B $\to$ Llama-3.1-70B}} \\
  \midrule
    PRM guided search & 65.4 & 71.8 & 77.3 & 40.2 & 43.4 & \textbf{46.8} & 11.0 & 13.2 & 12.5 & 18.3 & 17.5 & 20.0 & 0.0 & 2.2 & 5.6 \\
  \lgsshort\ (ours)             & 69.8 & 77.3 & 81.2 & 37.0 & 37.6 & 41.8 & 12.9 & \textbf{16.2} & 15.4 & 17.5 & 23.3 & 22.5 & \textbf{2.2} & 3.3 & 4.4 \\
  \cgsshort\ (ours)        & \textbf{71.3} & \textbf{79.3} & \textbf{83.9} & \textbf{42.4} & \textbf{45.0} & 46.2 & \textbf{16.2} & 15.1 & \textbf{21.7} & \textbf{25.0} & \textbf{25.0} & \textbf{29.2} & \textbf{2.2} & \textbf{6.7} &
  \textbf{10.0} \\
  \bottomrule
  \end{tabular}%
  }
  \caption{Accuracy (\%) for the matched-budget comparison with PRM guided search. PRM guided search scores naturally delimited reasoning steps using Qwen2.5-Math-PRM-72B. \lgsshort\ and \cgsshort\ score fixed-length chunks using the $L$ values chosen in Table~\ref{tab:l-selection}, with the corresponding large model in each pair (Qwen2.5-32B for the Qwen pair, Llama-3.1-70B for the Llama pair). AMC23 and AIME24 results are averaged over three independent random seeds.}
  \label{tab:main}
  \end{table*}

\section{Experiments}

Our experiments evaluate whether \methodname\ can improve small-model mathematical 
reasoning without training a reward model. We ask five questions:
(i) can it match PRM guidance under the same intervention budget 
\citep{zhang2025lessons,she2025rprm}?
(ii) why use fixed-length chunks rather than natural reasoning steps, and how 
sensitive are results to the choice of $L$?
(iii) how does it compare with post-hoc selection methods such as majority voting 
\citep{wang2023selfconsistency}, Best-of-$N$ \citep{cobbe2021gsm8k}, and 
self-certainty \citep{kang2025selfcertainty}?
(iv) do the gains persist when the small model is scaled from 1--1.5B to 7B parameters?
and (v) do the gains from \methodname\ come from generating longer reasoning traces?

\subsection{Experimental Setup}

\paragraph{Benchmarks and models.}
We evaluate on five mathematical reasoning benchmarks: GSM8K \citep{cobbe2021gsm8k} (1,319 questions), MATH \citep{hendrycks2021math} (500 questions), Minerva Math \citep{lewkowycz2022minerva} (272 questions), AMC23 (40 questions), and AIME24 (30 questions). Due to the small size of AMC23 and AIME24, we run all experiments on these two datasets with three different random seeds and report averages. Our main experiments use two small-to-large model pairs: Qwen2.5-1.5B $\to$ Qwen2.5-32B \citep{qwen25} and Llama-3.2-1B $\to$ Llama-3.1-70B \citep{llama3}. To test whether the gains persist with a stronger generator, we additionally evaluate Qwen2.5-7B $\to$ Qwen2.5-72B \citep{qwen25}. All models are served with vLLM \citep{kwon2023efficient}.

\paragraph{Chunk-level guidance.}
At each step, the small model samples $k \in \{8,16,32\}$ candidate chunks of $L$ tokens at temperature $0.7$. \lgsshort\ selects the highest-scoring chunk under Eq.~\ref{eq:guided}; \cgsshort\ uses Eq.~\ref{eq:contrastive}.

\paragraph{Prompting and evaluation.}
We prompt models using the template in Figure~\ref{app:fig:prompt} in the Appendix and extract the prediction from the \texttt{"answer"} field. Correctness is evaluated by first normalizing the prediction and reference answer: we remove \verb|\boxed{}| wrappers, standardize fraction notation, and strip units, currency symbols, commas, and whitespace. We then apply exact string matching, and also accept predictions that are numerically equivalent to the reference, such as \texttt{1/2} and \texttt{0.5}.

\subsection{Baselines}
We compare against baselines that differ in whether they use a scorer, which model provides the score, and when selection occurs.

\paragraph{\textbf{Majority@}$\mathbf{k}$~\citep{wang2023selfconsistency}} generates $k$ complete responses from the small model and selects the most frequent final answer.

\paragraph{\textbf{Self-Certainty} and \textbf{Borda count}~\citep{kang2025selfcertainty}}
rerank the $k$ completed responses using a distributional confidence score computed 
from a scorer model, without influencing generation.
For a response token sequence $y=(y_1,\ldots,y_n)$ scored under model $\pi$, 
Self-Certainty is
\[
    \mathrm{SC}_\pi(y)=-\frac{1}{nV}\sum_{i=1}^{n}\sum_{j=1}^{V}\log(V\cdot p_\pi(j\mid q,y_{<i})),
\]
where $n=|y|$, $V$ is the vocabulary size, $q$ is the input question, and $p_\pi(j\mid q,y_{<i})$ is the probability that model $\pi$ assigns to vocabulary token $j$ at position $i$; the score 
increases as the model's predicted distributions become more peaked.
Borda count aggregates by final answer: the response ranked $r$-th by $\mathrm{SC}_\pi$ 
contributes weight $(k-r+1)^p$ to its predicted answer, 
where $k$ is the number of candidates and $p$ controls how sharply top-ranked responses are weighted; the answer with the highest total weight is selected. We use $p{=}0.5$ in all experiments.
\citet{kang2025selfcertainty} use the generating model as the scorer ($\pi{=}\pi_s$); we additionally evaluate using the large model as the scorer ($\pi{=}\pi_l$), while keeping the small model as the generator in both cases. We refer to these as \textit{Self-Certainty (small)} / \textit{Borda count (small)} and \textit{Self-Certainty (large)} / \textit{Borda count (large)}, respectively.

\paragraph{\textbf{Best-of-$\mathbf{N}$}~\citep{cobbe2021gsm8k}} selects among the $k$ completed small-model responses using the guided score in Eq.~\ref{eq:guided}, with $L{=}2048$ chosen large enough that all responses finish before the limit. This is the full-response analogue of our method: the large model scores each completed response as a single unit, without mid-generation guidance.

\paragraph{\textbf{PRM guided search}~\citep{she2025rprm}} generates $k$ candidate continuations for the next reasoning step, scores each with a trained PRM, and appends the highest-scoring step before continuing generation. In our experiments, we use Qwen2.5-Math-PRM-72B~\citep{zhang2025lessons} as the PRM.

\begin{figure*}[h]
    \centering
    \includegraphics[width=0.8\textwidth]{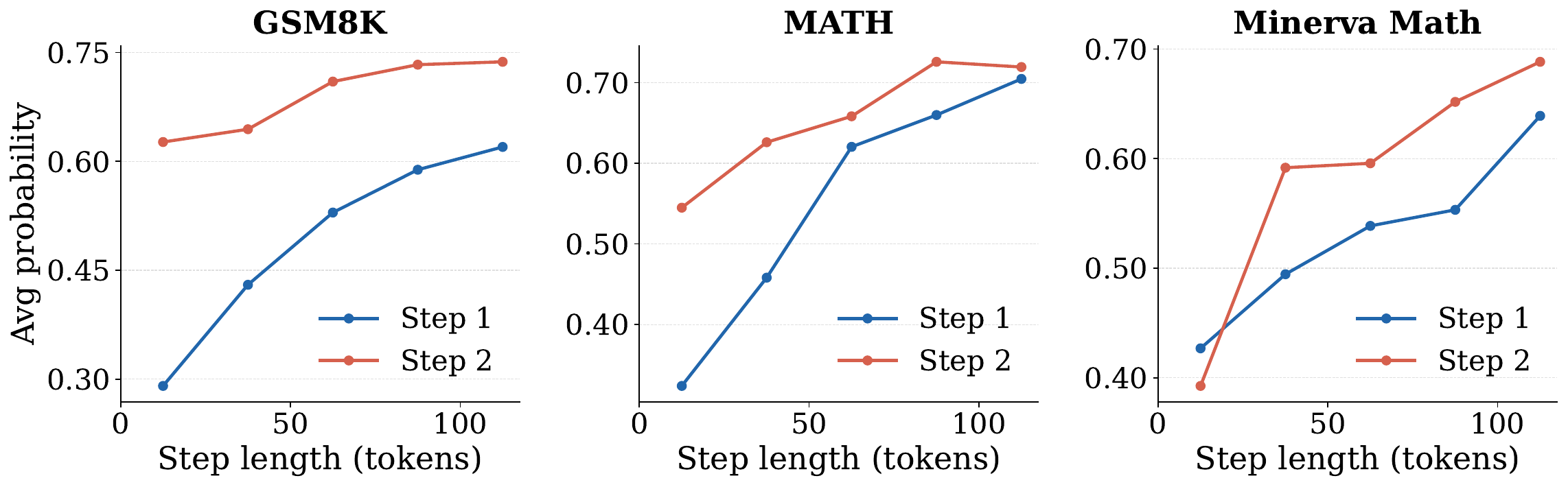}
    \caption{Length-normalized large-model probability vs.\ number of tokens in the first two reasoning steps for Qwen2.5-1.5B $\to$ Qwen2.5-32B, averaged over all examples in each dataset. Longer reasoning steps are assigned systematically higher probabilities by the large model even after length normalization, making variable-length step scoring unreliable for likelihood-based guidance.}
    \label{fig:step_logprob}
  \end{figure*}

  \begin{figure*}[t]
  \centering
  \includegraphics[width=0.8\textwidth]{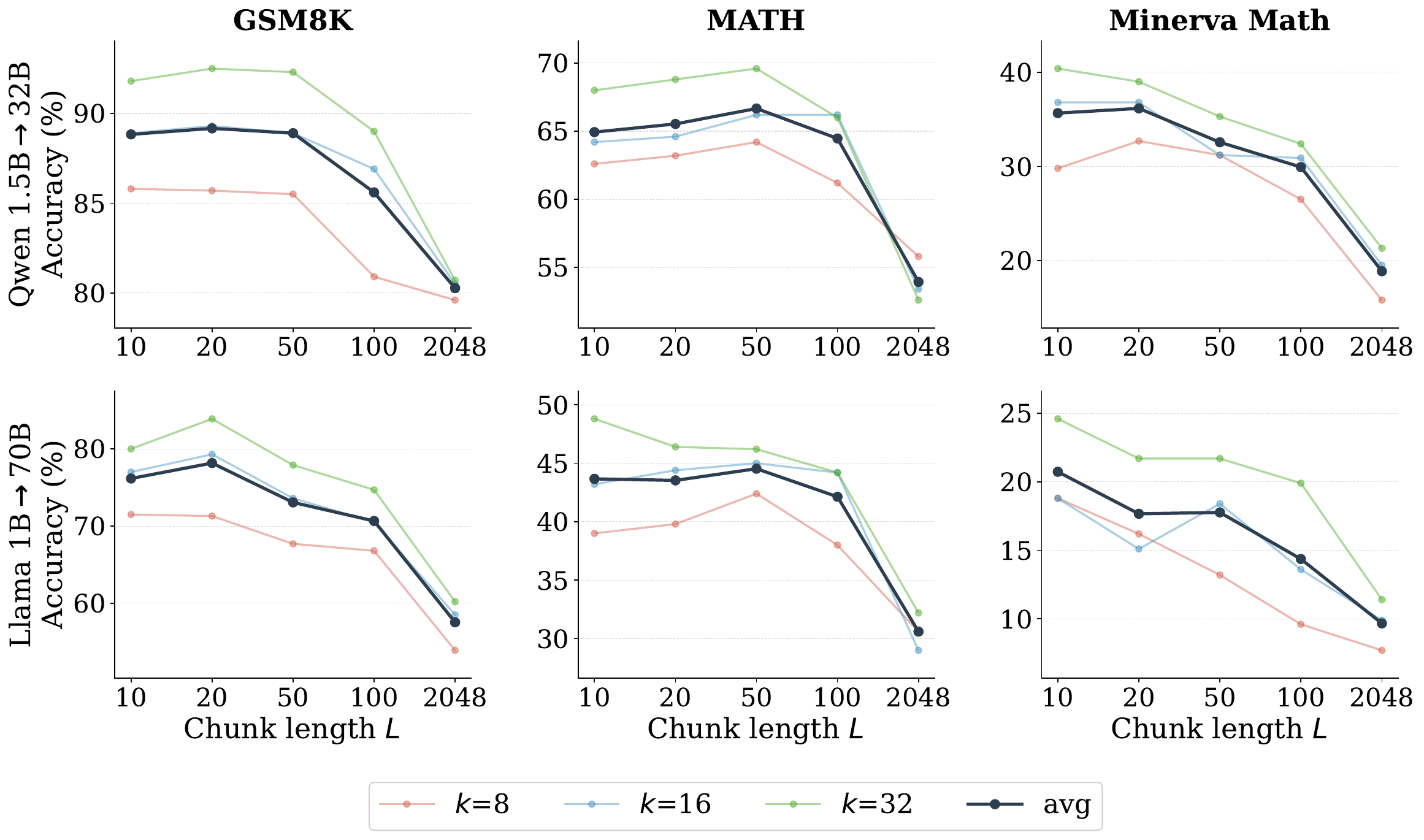}
\caption{\cgsshort\ accuracy vs.\ chunk length $L$ on GSM8K, MATH, and Minerva Math for both model pairs. The rightmost point ($L{=}2048$) corresponds to Best-of-$N$. Performance is stable for $L \in \{10, 20\}$ across all datasets; MATH remains stable through $L{=}50$. All datasets degrade sharply at $L{=}2048$.}
  \label{fig:l-sensitivity}
  \vspace{-6pt}
\end{figure*}

\subsection{Comparison with PRM Guided Search}

We compare our methods against PRM guided search under a matched guidance budget: for each dataset we select the chunk length $L \in \{10, 20, 50, 100\}$ whose average number of \cgsshort\ interventions per response most closely matches the average number of PRM scoring steps at $k{=}8$. Table~\ref{tab:l-selection} reports the chosen $L$ and the resulting intervention counts for both model pairs.

\begin{table}[h]
  \centering
  \small
  \setlength{\tabcolsep}{5pt}
  \begin{tabular}{l rrr rrr}
  \toprule
  & \multicolumn{3}{c}{\textbf{Qwen 1.5B$\to$32B}}
  & \multicolumn{3}{c}{\textbf{Llama 1B$\to$70B}} \\
  \cmidrule(lr){2-4}\cmidrule(lr){5-7}
  \textbf{Dataset} & $\boldsymbol{L}$ & \makecell[r]{\textbf{PRM}\\\textbf{interv.}} & \makecell[r]{\textbf{Chunk}\\\textbf{interv.}} & $\boldsymbol{L}$ & \makecell[r]{\textbf{PRM}\\\textbf{interv.}} &
  \makecell[r]{\textbf{Chunk}\\\textbf{interv.}} \\
  \midrule
  GSM8K   & 20 &  8.0 &  9.1 & 20 & 14.7 & 15.1 \\
  MATH    & 50 & 16.6 & 11.6 & 50 & 17.3 & 12.6 \\
  Minerva & 20 & 23.1 & 26.7 & 20 & 31.5 & 33.9 \\
  AMC23   & 50 & 23.2 & 17.9 & 50 & 24.3 & 17.4 \\
  AIME24  & 50 & 26.5 & 19.3 & 50 & 25.8 & 25.6 \\
  \bottomrule
  \end{tabular}
  \caption{Chosen chunk length $L$ and average number of large-model interventions per response at $k{=}8$. PRM interv.\ = average number of reasoning steps per response at which Qwen2.5-Math-PRM-72B scores
  candidates; Chunk interv.\ = average number of $L$-token chunks per response at which our method scores candidates. $L$ is selected as the value in $\{10,20,50,100\}$ whose chunk intervention count is closest to
  the PRM step count.}
  \label{tab:l-selection}
  \end{table}

Table~\ref{tab:main} reports accuracy under the matched guidance budgets from Table~\ref{tab:l-selection}.
For the Qwen pair, \cgsshort\ and PRM guided search perform comparably on GSM8K and MATH: the difference is at most 0.3~pp in either direction across $k \in \{8,16,32\}$. Without any reward-model training, \cgsshort\ outperforms PRM guided search on Minerva Math ($+$7.6~pp on average across $k$), AMC23 ($+$2.2~pp), and AIME24 ($+$1.1~pp). \lgsshort\ trails PRM on GSM8K and MATH, consistently outperforms it on Minerva Math, and is competitive on AMC23 and AIME24.

For the Llama pair, the gains are larger and more consistent.
\cgsshort\ outperforms PRM guided search on GSM8K, Minerva Math, AMC23, and AIME24 at every value of $k$, with average gains of $+$6.7~pp, $+$5.4~pp, $+$7.8~pp, and $+$3.7~pp, respectively; on MATH, \cgsshort\ leads at $k{=}8$ and $k{=}16$ but trails marginally at $k{=}32$ (46.2\% vs.\ 46.8\%), with an average gain of $+$1.1~pp.

Results on AMC23 ($n{=}40$) and AIME24 ($n{=}30$) are averaged over three random seeds to reduce variance from the small test sets. Overall, chunk-level likelihood guidance from an off-the-shelf large model can match or outperform PRM guided search at a comparable intervention budget, without training or using a reward model.

\subsection{Why Fixed-Length Chunks?}

A natural alternative to fixed-length chunks is to guide generation at the level of reasoning steps, as in PRM guided search. However, this requires comparing candidate continuations with variable lengths. To test this, we run a diagnostic analysis. We collect greedy small-model responses on GSM8K, MATH, and Minerva Math and split each response into reasoning steps. For each step, we score it with the large model conditioned on all preceding steps, using the same length-normalized log-probability as in Eq.~\ref{eq:guided}. We then plot the exponentiated score (average large-model probability) against step length in tokens for the first two reasoning steps, averaging over all examples in each dataset. Figure~\ref{fig:step_logprob} shows the result for Qwen2.5-1.5B $\to$ Qwen2.5-32B.

Across datasets and for both early reasoning-step positions, longer steps receive higher scores from the large model, even after length normalization. This creates a length bias: if we used naturally delimited reasoning steps as candidates for our method, the large model would tend to prefer longer continuations, not necessarily better ones. Fixed-length chunks avoid this confound: since all $k$ candidates at a given decision point have the same length $L$ and are conditioned on the same context, their scores are directly comparable.

\begin{table*}[h]
\centering
\caption{Accuracy (\%) for both model pairs; \lgsshort\ and \cgsshort\ use $L{=}20$. Bold = best non-oracle result per column. AMC23 and AIME24 results are averaged over three independent random seeds.}
\label{tab:results-l20}
\setlength{\tabcolsep}{3pt}
\resizebox{\textwidth}{!}{%
\begin{tabular}{l rrr rrr rrr rrr rrr}
\toprule
& \multicolumn{3}{c}{\textbf{GSM8K}} & \multicolumn{3}{c}{\textbf{MATH}} & \multicolumn{3}{c}{\textbf{Minerva}} & \multicolumn{3}{c}{\textbf{AMC23}} & \multicolumn{3}{c}{\textbf{AIME24}} \\
\cmidrule(lr){2-4}\cmidrule(lr){5-7}\cmidrule(lr){8-10}\cmidrule(lr){11-13}\cmidrule(lr){14-16}
\textbf{Method} & $k{=}8$ & $k{=}16$ & $k{=}32$ & $k{=}8$ & $k{=}16$ & $k{=}32$ & $k{=}8$ & $k{=}16$ & $k{=}32$ & $k{=}8$ & $k{=}16$ & $k{=}32$ & $k{=}8$ & $k{=}16$ & $k{=}32$ \\
\midrule
\multicolumn{16}{l}{\textit{Qwen2.5-1.5B $\to$ Qwen2.5-32B}} \\
\midrule
Small greedy    & 55.6 & 55.6 & 55.6 & 52.8 & 52.8 & 52.8 &  19.1 &  19.1 &  19.1 & 27.5 & 27.5 & 27.5 &  0.0 &  0.0 &  0.0 \\
Large greedy    & 94.8 & 94.8 & 94.8 & 78.0 & 78.0 & 78.0 & 50.0 & 50.0 & 50.0 & 52.5 & 52.5 & 52.5 & 16.7 & 16.7 & 16.7\\
Pass@$k$        & 88.6 & 92.7 & 95.1 & 71.0 & 78.6 & 83.8 & 38.2 & 50.7 & 57.4 & 56.7 & 63.3 & 70.0 &  7.8 & 16.7 & 25.6 \\
\midrule
Majority@$k$    & 74.0 & 77.7 & 79.7 & 56.6 & 59.6 & 63.0 & 21.3 & 24.3 & 26.1 & 28.3 & 36.7 & 34.2 &  2.2 &  3.3 &  4.4 \\
Self-Certainty (small) & 65.1 & 67.9 & 67.5 & 52.6 & 51.0 & 52.6 &  18.0 & 20.6 & 20.6 & 25.8 & 35.0 & 30.8 &  3.3 &  3.3 &  3.3 \\
Borda count (small)    & 75.0 & 78.0 & 80.1 & 57.0 & 59.6 & 63.2 & 22.1 & 25.7 & 26.1 & 35.0 & 35.8 & 37.5 &  3.3 &  4.4 &  3.3 \\
Self-Certainty (large) & 69.7 & 73.4 & 74.8 & 51.2 & 49.2 & 53.6 &  19.9 & 22.4 & 26.5 & 27.5 & 34.2 & 35.8 &  4.4 &  4.4 &  8.9 \\
Borda count (large)    & 75.2 & 78.7 & 81.0 & 57.6 & 60.0 & 63.4 & 22.8 & 26.8 & 28.3 & 32.5 & 36.7 & 38.3 &  4.4 &  5.6 &  5.6 \\
Best-of-$N$ & 79.6 & 80.5 & 80.7 & 55.8 & 53.4 & 52.6 & 15.8 & 19.5 & 21.3 & 25.0 & 28.3 & 25.0 & 3.3 & 0.0 & 3.3 \\
\midrule
PRM guided search         & \textbf{86.0} & 89.2 & 92.3 & \textbf{64.2} & \textbf{66.4} & \textbf{69.8} & 25.0 & 28.3 & 32.4 & 36.7 & 39.2 & 43.3 & 5.6 &  6.7 & 11.1 \\
\lgsshort\ (ours)          & 85.4 & 87.6 & 90.4 & 60.0 & 61.8 & 67.4 & 31.6 & 33.8 & 36.0 & 35.8 & 40.0 & 40.8 &  3.3 & \textbf{8.9} & 12.2 \\
\cgsshort\ (ours)     & 85.7 & \textbf{89.3} & \textbf{92.5} & 63.2 & 64.6 & 68.8 & \textbf{32.7} & \textbf{36.8} & \textbf{39.0} & \textbf{40.8} & \textbf{43.3} & \textbf{50.8} & \textbf{7.8} & \textbf{8.9} & \textbf{13.3} \\
\midrule
\multicolumn{16}{l}{\textit{Llama-3.2-1B $\to$ Llama-3.1-70B}} \\
\midrule
Small greedy    & 41.9 & 41.9 & 41.9 & 25.2 & 25.2 & 25.2 & 7.4 & 7.4 & 7.4 &  2.5 &  2.5 &  2.5 &  0.0 &  0.0 &  0.0 \\
Large greedy    & 95.1 & 95.1 & 95.1 & 68.0 & 68.0 & 68.0 & 46.0 & 46.0 & 46.0 & 45.0 & 45.0 & 45.0 & 20.0 & 20.0 & 20.0 \\
Pass@$k$        & 70.5 & 79.1 & 86.8 & 49.6 & 60.8 & 66.4 & 21.0 & 29.4 & 35.3 & 40.0 & 47.5 & 59.2 &  3.3 &  7.8 & 17.8 \\
\midrule
Majority@$k$    & 50.5 & 54.8 & 56.0 & 28.2 & 34.2 & 36.0 &  8.5 &  8.1 &  7.4 & 15.0 & 13.3 & 16.7 &  2.2 &  1.1 &  3.3 \\
Self-Certainty (small) & 41.6 & 44.7 & 46.6 & 23.0 & 24.4 & 25.8 &  8.8 &  7.0 &  9.2 & 11.7 & 12.5 & 18.3 & \textbf{3.3} &  0.0 &  1.1 \\
Borda count (small)    & 51.0 & 55.4 & 57.2 & 28.6 & 33.4 & 37.8 &  8.5 &  9.2 &  9.6 & 14.2 & 16.7 & 18.3 &  2.2 &  1.1 &  2.2 \\
Self-Certainty (large) & 55.3 & 59.2 & 62.7 & 29.8 & 30.0 & 33.4 &  9.6 &  9.6 &  11.8 & 14.2 & 15.8 & 19.2 &  1.1 &  3.3 &  2.2 \\
Borda count (large)    & 54.4 & 57.0 & 59.1 & 31.4 & 34.6 & 37.8 &  8.5 &  9.6 &  10.7 & 16.7 & 15.8 & 15.8 &  2.2 &  3.3 &  2.2 \\
Best-of-$N$ & 53.9 & 58.5 & 60.2 & 30.6 & 29.0 & 32.2 &  7.7 & 9.9 & 11.4 &  10.8 &  16.7 &  12.5 & 2.2 & 1.1 & 0.0 \\
\midrule
PRM guided search & 65.4 & 71.8 & 77.3 & \textbf{40.2} & 43.4 & \textbf{46.8} & 11.0 & 13.2 & 12.5 & 18.3 & 17.5 & 20.0 & 0.0 & 2.2 & 5.6 \\
\lgsshort\ (ours)          & 69.8 & 77.3 & 81.2 & 36.6 & 38.8 & 44.2 & 12.9 & \textbf{16.2} & 15.4 & \textbf{24.2} & 23.3 & \textbf{26.7} & \textbf{3.3} &  4.4 &  7.8 \\
\cgsshort\ (ours)     & \textbf{71.3} & \textbf{79.3} & \textbf{83.9} & 39.8 & \textbf{44.4} & 46.4 & \textbf{16.2} & 15.1 & \textbf{21.7} & 23.3 & \textbf{25.0} & 24.2 &  2.2 & \textbf{10.0} & \textbf{12.2} \\
\bottomrule
\end{tabular}%
}
\end{table*}

\subsection{Ablation on Chunk length $\boldsymbol{L}$}

Figure~\ref{fig:l-sensitivity} shows \cgsshort\ accuracy as a function of $L$ on GSM8K, MATH, and Minerva Math for both model pairs. For the Qwen pair, performance is stable for $L \in \{10, 20\}$ across all datasets; for the Llama pair, $L{=}10$ tends to perform best on Minerva Math while remaining comparable to $L{=}20$ elsewhere. On GSM8K and Minerva Math, accuracy begins to decline at $L{=}50$; MATH is less sensitive and peaks at $L{=}50$. Performance degrades further at $L{=}100$ and sharply at $L{=}2048$, where the chunk is set large enough to cover any complete response and the large model scores only the finished output, with no opportunity to intervene during generation. We use $L{=}20$ for all remaining experiments, as it lies in the stable region across all datasets and model pairs without requiring per-dataset tuning.

\subsection{Main Results}

Table~\ref{tab:results-l20} compares all methods for the two small-model pairs (Qwen2.5-1.5B$\to$32B and Llama-3.2-1B$\to$70B): \lgsshort\ (ours) and \cgsshort\ (ours) at $L{=}20$, PRM guided search at natural reasoning-step boundaries, and post-hoc methods (Majority@$k$, Self-Certainty, Borda count, Best-of-$N$) on complete responses.

\begin{table*}[t]
  \centering
  \footnotesize
  \setlength{\tabcolsep}{3pt}
  \begin{tabular}{l r r r r r r}
  \toprule
  \textbf{Method} & \textbf{GSM8K} & \textbf{MATH} & \textbf{Minerva} & \textbf{AMC23} & \textbf{AIME24} & \textbf{Avg} \\
  \midrule
  Small greedy              & 82.6 & 70.0 & 51.8 & 50.0 & 10.0 & 52.9 \\
  Large greedy              & 90.5 & 80.6 & 64.3 & 65.0 & 20.0 & 64.1 \\
  Pass@$k$                  & 97.1 & 90.8 & 80.1 & 84.2 & 28.9 & 76.2 \\
  \midrule
  Majority@$k$              & 91.9 & 77.2 & 57.7 & 59.2 & 14.4 & 60.1 \\
  Self-Certainty (small)    & 88.2 & 72.8 & 53.3 & 47.5 & 12.2 & 54.8 \\
  Borda count (small)       & 89.8 & 78.0 & 56.6 & 60.0 & 15.6 & 60.0 \\
  Self-Certainty (large)    & 90.8 & 75.8 & 55.1 & 49.2 & 15.6 & 57.3 \\
  Borda count (large)       & 90.1 & 77.4 & 58.5 & 58.3 & 15.6 & 60.0 \\
  Best-of-$N$               & 92.0 & 75.8 & 48.9 & 50.8 & 11.1 & 55.7 \\
  \midrule
  PRM-72B guided search     & \textbf{95.5} & \textbf{84.0} & \textbf{63.6} & 65.0 & \textbf{17.8} & \textbf{65.2} \\
  \lgsshort\ (ours)         & 92.1 & 80.2 & 62.5 & 60.0 & 16.7 & 62.3 \\
  \cgsshort\ (ours)         & 91.7 & 81.8 & \textbf{63.6} & \textbf{65.8} & 15.6 & 63.7 \\
  \bottomrule
  \end{tabular}
  \caption{Accuracy (\%) for Qwen2.5-7B $\to$ Qwen2.5-72B at $k{=}16$, $L{=}20$. Bold indicates the best non-oracle result per column. AMC23 and AIME24 results are averaged over three independent random seeds.}
  \label{tab:qwen7b-k16}
\end{table*}

\textbf{Post-hoc selection fails.}

Best-of-$N$ shows mixed results: it outperforms Majority@$k$ on GSM8K 
across both model pairs at all values of $k$, where large-model scoring 
provides a reliable signal. On MATH, Best-of-$N$ falls below 
Majority@$k$ for the Qwen pair at all values of $k$ and for the Llama 
pair at $k{=}16$ and $k{=}32$. On Minerva Math and AMC23, results are 
mixed: Best-of-$N$ consistently underperforms Majority@$k$ for the Qwen 
pair, while the Llama pair shows no clear pattern. Moreover, Best-of-$N$ can degrade as $k$ grows: on Qwen MATH it 
drops from 55.8\% at $k{=}8$ to 52.6\% at $k{=}32$, while Majority@$k$ 
improves from 56.6\% to 63.0\% over the same range.

Self-Certainty (small) underperforms Majority@$k$ in 24 out of 30 settings, showing that small-model confidence is not a reliable signal for picking the best individual response, consistent with the findings of~\citet{kang2025selfcertainty}. Borda count (small) recovers some of this by falling back on the voting mechanism, offering marginal gains over Majority@$k$ across most settings.
Using the large model as the confidence scorer, Borda count (large) generally improves over Borda count (small), as does Self-Certainty (large) over Self-Certainty (small); Borda count (large) also outperforms Majority@$k$ in nearly all settings, with larger margins on the Llama pair. Post-hoc methods nonetheless remain substantially below chunk-level guidance in nearly all settings.

\textbf{Pass@$k$ as the oracle ceiling.}
Pass@$k$ measures whether the correct answer appears in any of the $k$ independent small-model samples, setting the oracle ceiling for post-hoc selection methods that choose among completed responses. On the Qwen pair at $k{=}32$, this ceiling is 95.1\% on GSM8K and 57.4\% on Minerva Math; no post-hoc method approaches it. Chunk-level guidance is not bound by this ceiling: by intervening during generation rather than after it, the large model can steer the small model toward solutions that no unguided sample would have reached. In three settings chunk-level guidance surpasses this ceiling: 
\cgsshort\ on the Llama pair exceeds Pass@$k$ on GSM8K at $k{=}8$ 
(71.3\% vs.\ 70.5\%) and $k{=}16$ (79.3\% vs.\ 79.1\%), and on 
AIME24 at $k{=}16$ (10.0\% vs.\ 7.8\%).

\textbf{Chunk-level guidance substantially improves reasoning.}
Both \lgsshort\ and \cgsshort\ substantially outperform all post-hoc methods. On the Qwen pair at $k{=}32$, \cgsshort\ reaches 92.5\% on GSM8K and 50.8\% on AMC23, gains of $+$12.8 and $+$16.6~pp over Majority@$k$. On the Llama pair at $k{=}32$ the gains are even larger: $+$27.9~pp on GSM8K (83.9\%) and $+$14.3~pp on Minerva Math (21.7\%).

\textbf{\cgs\ outperforms \lgs.}
\cgsshort\ outperforms \lgsshort\ scoring in most settings. Rather than measuring how likely the large model finds a chunk in absolute terms, \cgsshort\ measures how much more likely the large model finds it relative to the small model, capturing where the large model's guidance adds the most value. On average, \cgsshort\ outperforms \lgsshort\ by 2.8~pp on the Qwen pair and 2.2~pp on the Llama pair across all datasets and values of $k$.

\textbf{Comparison with PRM at $\boldsymbol{L{=}20}$.}
\cgsshort\ outperforms PRM guided search on Minerva Math, AMC23, and
  AIME24 across both model pairs and all values of $k$.
  On GSM8K, \cgsshort\ outperforms PRM on the Llama pair but falls
  marginally behind on the Qwen pair at $k{=}8$ (85.7\% vs.\ 86.0\%).
  MATH is the only dataset where PRM has a consistent edge: \cgsshort\ 
  falls below PRM on the Qwen pair across all $k$ (e.g., 68.8\% vs.\
  69.8\% at $k{=}32$) and on the Llama pair at $k{=}8$ and $k{=}32$.
Averaged across all five datasets and three values of $k$, \cgsshort\ 
outperforms PRM by $+$2.8~pp on the Qwen pair and $+$4.7~pp on the 
Llama pair.

\subsection{Scaling to a Larger Small Model: Qwen2.5-7B $\to$ 72B}
To test whether the gains hold when the small model is already competitive, we run the same setup with Qwen2.5-7B guided by Qwen2.5-72B at $k{=}16$, $L{=}20$. Table~\ref{tab:qwen7b-k16} reports results across all five datasets. \cgsshort\ achieves an average accuracy of 63.7\%, outperforming the best post-hoc selection method by $+$3.6~pp on average and nearly matching the 72B greedy ceiling (64.1\%), surpassing it on GSM8K (91.7\% vs.\ 90.5\%), MATH (81.8\% vs.\ 80.6\%), and AMC23 (65.8\% vs.\ 65.0\%), confirming that chunk-level guidance remains effective at this scale. On average, \cgsshort\ trails PRM-72B guided search by $1.5$~pp; on Minerva Math it ties PRM (63.6\%) and on AMC23 it surpasses it (65.8\% vs.\ 65.0\%); the largest gap to PRM is on GSM8K (91.7\% vs.\ 95.5\%). The smaller gap between the two models in this setting --- 11.2~pp between 7B and 72B greedy, compared to 27.4~pp between 1.5B and 32B (Table~\ref{tab:results-l20}) --- leaves less room for likelihood-based scoring to differentiate among candidates, which may explain why chunk-level guidance falls short of a dedicated reward model in this setting, unlike the smaller model pairs in Table~\ref{tab:results-l20}.

\subsection{Analysis of Reasoning Trace Length}
\label{sec:response-length}
Because longer reasoning traces can improve mathematical reasoning by providing more intermediate computation, we ask whether the gains from \methodname\ are explained by longer generations. Table~\ref{tab:response-length} suggests the opposite. We measure reasoning trace length as the number of tokens before the final answer. We compare greedy small-model and greedy large-model generation with PRM guided search and \cgsshort\ on the Qwen2.5-1.5B $\to$ Qwen2.5-32B pair; the two guided methods are evaluated at $k{=}8$, with \cgsshort\ using $L{=}20$.

\begin{table}[h]
    \centering
    \small
    \setlength{\tabcolsep}{4pt}
    \begin{tabular}{l rrrr r}
    \toprule
    \textbf{Dataset} 
    & \makecell[r]{\textbf{Greedy}\\\textbf{small}} 
    & \makecell[r]{\textbf{Greedy}\\\textbf{large}} 
    & \textbf{PRM} 
    & \textbf{\cgsshort} 
    & \makecell[r]{\textbf{PRM}/\\\textbf{\cgsshort}} \\
    \midrule
    GSM8K   &  159 &  164 & 311 & 173 & $\times$1.80 \\
    MATH    &  652 &  385 & 589 & 492 & $\times$1.20 \\
    Minerva &  651 &  464 & 773 & 524 & $\times$1.48 \\
    AMC23   &  906 &  618 & 911 & 737 & $\times$1.24 \\
    AIME24  & 1091 &  729 & 925 & 884 & $\times$1.05 \\
    \bottomrule
    \end{tabular}
    \caption{Average reasoning trace length in tokens on Qwen2.5-1.5B $\to$ Qwen2.5-32B. Greedy baselines use single-sample generation, while PRM guided search and \cgsshort\ are evaluated at $k{=}8$; \cgsshort\ uses $L{=}20$. The final column reports the PRM/\cgsshort\ trace-length ratio.}
\label{tab:response-length}
\end{table}

The large model's greedy responses are generally shorter than the small model's, especially on harder datasets such as MATH (385 vs.\ 652 tokens) and AIME24 (729 vs.\ 1{,}091 tokens). This suggests that the stronger model often reaches solutions through more direct reasoning paths. Since \cgsshort\ guides the small model using the large model's likelihoods, it partially inherits this conciseness: compared with the small-model greedy baseline, \cgsshort\ produces shorter traces on MATH, Minerva, AMC23, and AIME24, moving closer to the large model's reasoning length.

PRM guided search does not show the same pattern. On GSM8K and Minerva, PRM produces longer traces than the small-model greedy baseline (311 vs.\ 159 tokens on GSM8K, and 773 vs.\ 651 on Minerva). Directly comparing the two guided methods, \cgsshort\ produces shorter reasoning traces than PRM guided search on all five datasets. The gap is largest on GSM8K, where PRM responses contain 80\% more reasoning tokens than \cgsshort\ responses (311 vs.\ 173 tokens). Thus, the gains from \methodname\ do not come from simply generating longer reasoning traces; instead, large-model likelihood guidance appears to steer the small model toward more concise reasoning paths. This is consistent with prior findings that longer reasoning traces do not necessarily translate into better reasoning performance. \citep{hassid2025don,wu2025more,zhou2026more, chegini2025reasoning}.

\section{Conclusion}
We presented \methodname, a training-free method for improving small-model mathematical reasoning by using a large language model to score fixed-length candidate chunks during generation.
By intervening at each chunk step rather than only after generation is complete, our method can steer the small model before incorrect reasoning paths fully develop, addressing a core limitation of post-hoc selection.

Our main finding is that an off-the-shelf large language model can be a surprisingly effective scorer. Across Qwen2.5-1.5B $\to$ 32B and 
Llama-3.2-1B $\to$ 70B on GSM8K, MATH, Minerva Math, AMC23, and 
AIME24, chunk-level guidance substantially outperforms majority voting, Best-of-$N$, and self-certainty-based reranking, while matching or outperforming PRM guided search on most benchmarks without any reward-model training. \cgsshort, which rewards chunks where the large model's preference diverges from the small model's, provides additional gains in most settings. A key design insight is that fixed-length chunks are necessary: large-model 
log-probabilities are systematically biased toward longer reasoning 
steps, making variable-length step scoring unreliable.

\section*{Limitations}

All experiments are on mathematical reasoning benchmarks; whether chunk-level likelihood guidance generalizes to other domains such as coding or commonsense reasoning remains an open question.
We evaluate only within-family model pairs (Qwen$\to$Qwen and Llama$\to$Llama), so the effectiveness of cross-family pairs is untested.
Finally, our length-bias analysis motivates fixed-length chunks, but whether a better-calibrated variable-length scoring scheme could match our approach remains unexplored.

\section*{Acknowledgments}

This project was supported in part by a grant from an NSF CAREER AWARD 1942230, the ONR PECASE grant N00014-25-1-2378, ARO’s Early Career Program Award 310902-00001, Army Grant No. W911NF2120076, the NSF award CCF2212458, NSF Award No. 2229885 (NSF Institute for Trustworthy AI in Law and Society, TRAILS), a MURI grant 14262683, DARPA AIQ grant HR00112590066, an award from Google Research and an award from meta 314593-00001.

\bibliography{main}

\appendix
\section{Prompt Template}

\begin{figure}[h]
\centering
\small
\begin{tabular}{|p{0.93\columnwidth}|}
\hline
\\[-4pt]
Solve this math problem step by step, then provide your answer.\\[4pt]
\texttt{[PROBLEM]}\\[4pt]
Think through the problem step by step, showing your reasoning. Then provide your final answer in this exact format:\\[4pt]
\texttt{\{}\\
\hspace*{1em}\texttt{"reasoning": "your step-by-step solution here",}\\
\hspace*{1em}\texttt{"answer": "your\_final\_answer"}\\
\texttt{\}}\\[4pt]
Put only the final answer in the \texttt{"answer"} field. Use LaTeX formatting if needed (e.g., \texttt{\textbackslash frac\{a\}\{b\}} for fractions).\\[2pt]
\hline
\end{tabular}
\caption{Instruction template used for all datasets. The model is prompted to reason step by step~\citep{wei2022chain} and produce a structured JSON response; we extract the predicted answer from the \texttt{"answer"} field.}
\label{app:fig:prompt}
\end{figure}

\end{document}